\documentclass[a4paper, 10pt, conference]{ieeeconf}      
\usepackage{FG2019}

\usepackage{graphicx}
\usepackage{caption}
\usepackage{subcaption}
\FGfinalcopy 

\IEEEoverridecommandlockouts                              
\def\FGPaperID{177} 

\title{\LARGE \bf Active Authentication using an Autoencoder regularized CNN-based One-Class Classifier 
}

\author{\parbox{16cm}{\centering
    {\large Poojan Oza, Vishal M. Patel}\\
    {\normalsize
    Department of Electrical and Computer Engineering, \\Johns Hopkins University, 3400 N. Charles St, Baltimore, MD 21218, USA\\}
    {\normalsize
     \textit{email: \{poza2, vpatel36\}@jhu.edu}\\}}
     \thanks{P. Oza and V. M. Patel are with the Department of Electrical and Computer Engineering, Johns Hopkins University, Baltimore, MD, USA. Email: \{poza2, vpatel36\}@jhu.edu.
     	This work was supported by the NSF grant 1801435.}
}

\begin{document}

\ifFGfinal
\thispagestyle{empty}
\pagestyle{empty}
\else
\author{Anonymous FG 2019 submission\\ Paper ID \FGPaperID \\}
\pagestyle{plain}
\fi
\maketitle

\begin{abstract}
Active authentication refers to the process in which users are unobtrusively monitored and authenticated continuously throughout their interactions with mobile devices.  Generally, an active authentication problem is modelled as a one class classification problem due to the unavailability of data from the impostor users.  Normally, the enrolled  user is considered as the target class (genuine) and the unauthorized users are considered as unknown classes (impostor).  We propose a convolutional neural network (CNN) based approach for one class classification in which a zero centered Gaussian noise and an autoencoder  are used to model the pseudo-negative class and to regularize the network  to learn meaningful  feature representations for one class data, respectively.  The overall network is trained using a combination of the cross-entropy and the reconstruction error losses.  A key feature of the proposed approach is that any pre-trained CNN can be used as the base network for one class classification.  Effectiveness of  the  proposed  framework  is  demonstrated  using  three publically available face-based active authentication datasets and it is shown that the proposed method achieves superior performance compared to the traditional one class classification methods. The source code is available at : github.com/otkupjnoz/oc-acnn.
\end{abstract}

\section{INTRODUCTION}

Biometric information based on physical characteristics such as  face, iris, fingerprints etc., have been an integral part of security (e.g., airport security, law enforcement) and authentication systems (e.g., banking transactions). Physiological biometrics are never lost or forgotten and are difficult to forge, which gives such systems an edge over the traditional passcode based  approaches \cite{prabhakar2003biometric}. Moreover, these methods are convenient, user-friendly and secure compared to the traditional  explicit  authentication mechanisms.  Due to these reasons, biometric-based mobile authentication systems are receiving increased attention in recent years.

Since the introduction of the first fingerprint authentication-based mobile device in 2013, the smartphone industry has introduced a variety of mobile devices that make use of biometric authentication.  According to a recently published study, by the year 2020 almost 100\% of the smartphones will use biometric authentication as a standard feature \cite{cmaxine2016casestudy}. Furthermore, recent years have witnessed a major shift, where smartphone makers have started moving away from password-based or fingerprint-based systems to face-based authentication systems. Considering the advances in face recognition, it won't be too long before face-based authentication becomes a norm for mobile devices.

However, face-based verification systems are explicit in nature and are  fundamentally limited.  For instance,  as  long  as  the  mobile  phone  remains  active,  typical  devices  incorporate  no  mechanisms
to  verify  that  the  user  originally  authenticated  is  still  the  user  in  control  of  the  mobile  device.  Thus,
unauthorized individuals may improperly obtain access to personal information of the user if the system is spoofed or if the user does not exercise adequate vigilance after initial authentication.  To deal with this issue, Active Authentication (AA) systems were introduced in which  users  are  continuously  monitored  after
the  initial  access  to  the  mobile  device \cite{guidorizzi2013security}.  Various methods have been proposed for AA including touch gesture-based \cite{pozo2017exploring}, \cite{zhang2015touch}, \cite{perera2017extreme}, \cite{perera2017towards}, gait pattern-based \cite{derawi2010unobtrusive} and face-based \cite{fathy2015face}, \cite{crouse2015continuous}, \cite{samangouei2016convolutional}, \cite{perera2018efficient}, \cite{perera2016quickest} systems.  In particular, face-based AA systems have gained a lot of attraction in recent years.

\begin{figure}[t!]
	\centering
	\includegraphics[scale=0.80]{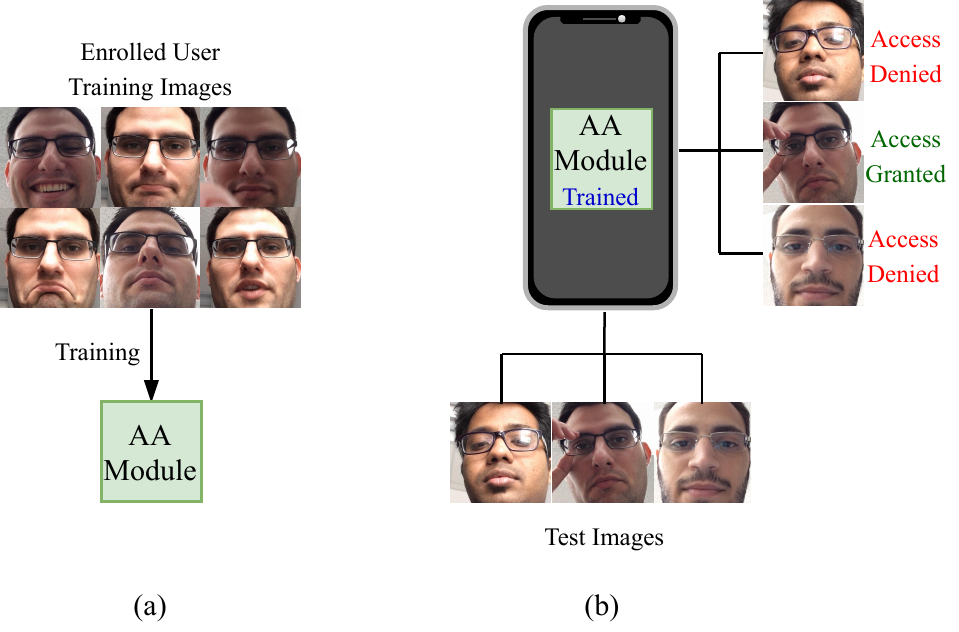}
	\caption{An overview of a typical AA system. (a) Data corresponding to the enrolled user are used to train an AA system. (b) During testing, data corresponding to the enrolled user as well as unknown user may be presented to the system.  The AA system then grants access to the enrolled user and blocks access to unknown users.}
	\label{fig:aa_sys}
\end{figure}

A naive approach for face-based AA would be to use face images correspoinding to all users and train a system to classify each user in a multi-class fashion. However, such an approach becomes counter-intuitive for AA since it requires the storage of all face images at a centralized location, raising data privacy issues \cite{patel2016continuous}. Hence, one must consider only the data collected from the enrolled user to develop an AA system. In other words, we need to explore possibilities of implementing AA systems using only the user's enrolled data.  This motivates us to view AA as a one class classification problem \cite{antal2015evaluation}. Fig. \ref{fig:aa_sys} shows a typical face-based AA system, modelled as a one class classification problem.  For the rest of this paper, we term the enrolled user and unauthorized user data as the target class and unknown class data, respectively.

\begin{figure*}[!t]
	\centering
	\includegraphics[scale=0.90]{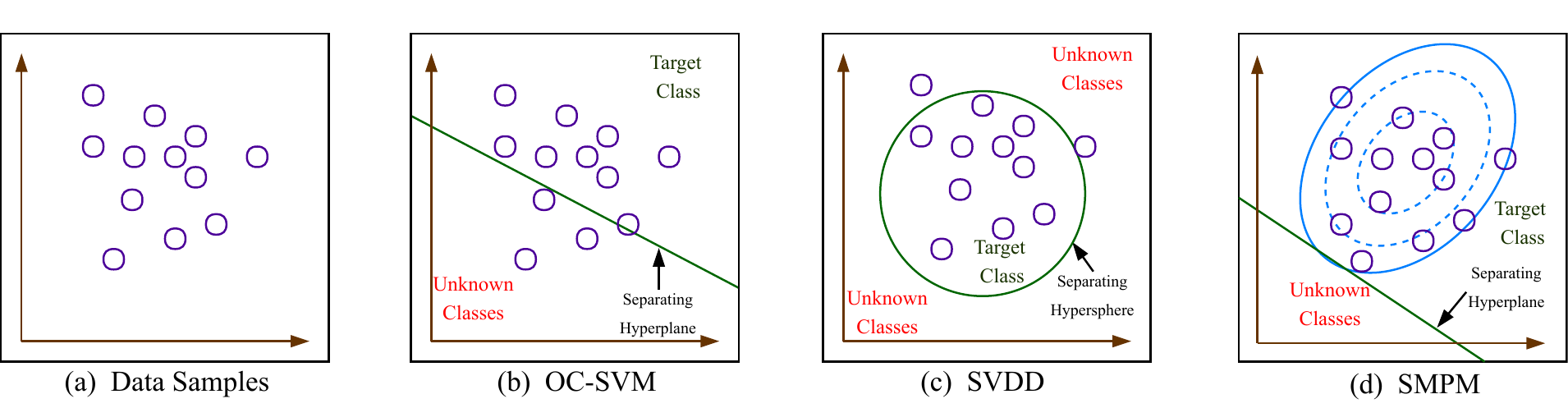}
	\caption{A graphical illustration of popular statistical one-class classification methods. (a) Sample data corresponding to a target class. (b) One-Class Support Vector Machines (OC-SVM), learn a hyperplane separating the data by maximizing the margin against the origin. (c) Support Vector Data Description (SVDD), learns a hypersphere in the feature space that encloses the given one-class data. (d) Single Minimax Probability Machines (SMPM), learn a hyperplane that minimizes the misclassification probability.}
	\label{fig:stat_methods}
\end{figure*}

Learning a one class classifier based on only the target class data  has been one of the most challenging problems in machine learning. Some of the earlier works have used statistical methods to tackle this problem. Such statistical methods usually seek separating hyperplane/hypersphere in the feature space to enclose the target class data \cite{scholkopf2001estimating}, \cite{tax2004support}, \cite{ghaoui2003robust}, \cite{khan2014one}. These methods rely on the quality of the representations used for the target class data. Earlier approaches were based on the hand-crafted features. In recent years features based on Deep Convolutional Neural Networks (CNNs) have shown to produce better results than hand-crafted features.  Several one class approaches have also been proposed in the literature that instead of learning a decision boundary, try to leverage CNNs to learn representations for the target class data. Most of these approaches are based on generative methods such as Auto-Encoders and Generative Adversarial Networks (GANs) \cite{goodfellow2014generative}. These generative approaches either use reconstruction errors or discriminator scores to identify the target class data \cite{sabokrou2018adversarially}, \cite{sabokrou2018deep}, \cite{lawson2017finding}, \cite{ravanbakhsh2017abnormal}, \cite{zhou2017anomaly}. However, features learned using these generative approaches are not discriminative enough compared to the features learned in traditional discriminative fashion. Attempts have been made to combine deep features with the statistical classifiers for learning better representations for one class classification \cite{erfani2016high}, \cite{andrews2016transfer}. Though, utilizing these powerful feature representations help in learning good decision boundaries, feature representations and classifiers are learned separately. In such a disjoint approach, classification module doesn't influence CNNs to modify the feature representation for a given target class data.  Several recent works have explored joint learning of both features and classifiers \cite{perera2018learning}, \cite{chalapathy2018anomaly} for one class classification. These methods demonstrated that representation learning together with classifier training results in improved performance. Based on this motivation, an end-to-end learning approach is proposed in this paper which jointly learns feature representations and a classifier for one class classification. Furthermore, the learned representations are constrained by a decoder network which regularizes the learned representations by enforcing them to reconstruct the original data. In summary, this paper makes the following contributions:

\begin{enumerate}
\item[$\bullet$] A new method is proposed for jointly learning representations and the decision boundary for one class classification.
\item[$\bullet$] A key feature of the proposed approach is that any  pre-trained CNN architecture (i.e. AlexNet, VGG16 and VGGFace) can be used as the base network.  
\item[$\bullet$] Extensive experiments are conducted on three face-based AA datasets and it is shown that the proposed approach can outperform many other statistical and deep learning-based one class classification approaches.
\end{enumerate}

\label{sec:introduction}

\section{Related Work}

Many statistical approaches have been explored over the years to address the one class classification problem. A graphical illustration of popular statistical one-class classification methods is given in Fig. \ref{fig:stat_methods}. One class support vector machines (OC-SVM), introduced by Scholkopf et al. \cite{scholkopf2001estimating}, learns a  classification boundary in the feature space by maximizing the margin of separating hyperplane against the origin. Fig. \ref{fig:stat_methods}(b) shows a typical example of a hyperplane learned by the OC-SVM. Another popular class of approaches for one class classification are based on Support Vector Data Description (SVDD), proposed by Tax et al. \cite{tax2004support}. As shown in the Fig. \ref{fig:stat_methods}(c), SVDD learns a separating hypersphere such that it encloses the majority of target class data. Another popular one class classification method is Single-MiniMax Probability Machines (SMPM) \cite{ghaoui2003robust} which is essentailly based on Minimax Probability Machines (MPMs) \cite{lanckriet2002minimax}. SMPM, similar to OC-SVM learns a hyperplane by maximizing margin against the origin. SMPM also considers the second order statistics of the data into its formulation which provides additional information on the structure of the data. This helps SMPM to find the best direction for separating the hyperplane as shown in Fig. \ref{fig:stat_methods}(d). Many variations of these methods have also been explored in the literature.  We refer readers to \cite{khan2014one} for a survey of different one-class classification methods.

As disucssed earlier, most of these classifiers typically use feature representations extracted from the data for classification. Hence, the quality of these feature representations become a crucial aspect of classifier learning and can affect the classification performance significantly.  As a result,  representation learning-based methods for one class classification have also been proposed in the literature. These methods leverage CNNs to learn representations for a given target class data. These methods often make use of the generative networks such as auto-encoders or GANs in which reconstruction errors or discriminator scores are utilized to identify the target class data \cite{sabokrou2018adversarially}, \cite{sabokrou2018deep}, \cite{lawson2017finding}, \cite{ravanbakhsh2017abnormal}, \cite{zhou2017anomaly}. However, features learned using these generative approaches are not as powerful as those learned in a discriminative fashion. However, training methodology used for discriminative feature learning requires data from multiple classes, limiting its application to multi-class classification. Several works have utilized features from pre-trained CNNs together with  one of the statistical approaches such as OC-SVM, SVDD, or SMPM to learn the decision boundary for one class classification \cite{erfani2016high}, \cite{andrews2016transfer}. Though, utilizing these powerful feature representations help in learning good decision boundaries, representations and decision boundaries are learned seperately. In such approaches, classification module doesn't influence CNNs for improving the feature representations. Classifier and representations, if jointly learned, have shown to improve the performance of the system as a whole \cite{perera2018learning}, \cite{chalapathy2018anomaly}, \cite{oza2019one}.

Perera et al. \cite{perera2018learning} proposed a one class algorithm to learn both features and classifier together. However, their approach requires an external dataset and can work only when multiple class data are available, making it not useful for AA systems. Chalpahty et al. \cite{chalapathy2018anomaly} introduced a neural network-based one class approach called One Class classification with Neural Networks (OC-NN) which jointly learns classifier and feature representation without using any external data. OC-NN introduced a novel loss function, designed for one class classification. This loss function enables classifier to influence the representation learning process. Similar to OC-NN, the proposed approach aims to jointly learn classifier and feature representation. However, the proposed approach is completely different from that proposed in OC-NN.  The feature extractor and the classifier are trained by introducing a Gaussian vector in the feature space which acts as a pseudo-negative class. Oza et al. \cite{oza2019one} showed that providing pseudo negative information helps learning better representations which often results in improved classification performance. In this paper, we introduce a decoder network in the pipeline which essentially constraints the feature extractor to generate feature representation that can reconstruct the original data. These self-representation constraint acts as a regularizer and helps improving the learned representation. We train the entire  networks in an end-to-end fashion using only the enrolled user data. In what follows, we present details of the proposed one class classification method based on autoencoder regularized CNN (OC-ACNN).

\label{sec:related_work}

\section{Proposed Approach}

\begin{figure*}[htp!]
	\centering
	\includegraphics[scale=1.250]{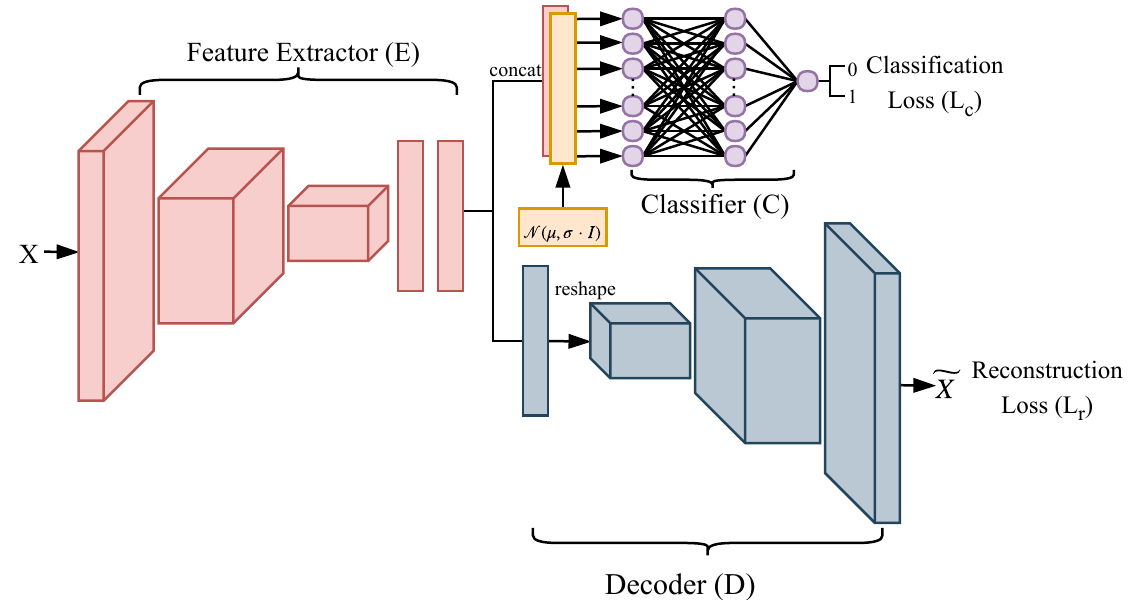}
	\caption{An overview of the  proposed OC-ACNN method. Here, $X$ is the input. The feature extractor module ($E$) can be any pre-trained CNN architecture. In this paper AlexNet, VGG16 and VGGFace networks are explored. The decoder module ($D$) is a simple four layer fully convolutional network. The decoder network essentially reconstructs the input image $X$, as $\tilde{X}$. The classification network ($C$ ) is a fully connected neural network trained to distinguish between feature vectors coming from $E$ and the Gaussian vectors sampled from $\mathcal{N}(\mu, \sigma \cdot I)$. The entire network is trained using a combination of classification loss ($\mathcal{L}_c$) and reconstruction loss ($\mathcal{L}_r$).}
	\label{fig:prop_app}
\end{figure*}

An overview of the proposed OC-ACNN network architecture  is shown in Fig. \ref{fig:prop_app}. It consists of three major modules namely, feature extractor network, classification network and decoder network.  The feature extractor network generates latent space representations for a given target class data. These latent representations are then fed to a classifier and a decoder network. Before feeding them to the classifier network, they are  concatenated with a vector sampled from a zero centered Gaussian  $\mathcal{N} (\mu, \sigma \cdot I)$, where $\sigma$ and $\mu$ are the parameters of the Gaussian and $I$ is the identity matrix. This Gaussian vector acts as a pseudo-negative class for the classifier. The classifier network is tasked with discriminating the target class representation from the pseudo-negative Gaussian vector. The decoder network takes in the same latent representation to reconstruct the original input. This enforces the latent representation generated by the feature extractor network to be self-representative i.e., representations are required to generate back the original input images. The classification network and the decoder network are trained end-to-end using a combination of binary cross entropy loss and $L1$ loss, respectively. 

\subsection{Feature Extractor}
The feature extractor network ($E$) can be any state of the art network architecture. In this paper, pre-trained AlexNet \cite{krizhevsky2012imagenet}, VGG16 \cite{simonyan2014very} and VGGFace \cite{parkhi2015deep} are considered. Before using these architectures as feature extractor, the final layer (i.e. softmax regression layer) is removed. While training, we update weights of only the fully connected layers and freeze the weights of convolutional layers. AlexNet and VGG16 utilized here are initialized with the ImageNet pre-trained weights and VGGFace is initialized with the VGGFace dataset pre-trained weights.

\subsection{Classification Network}
Assuming that the extracted features are $D$-dimensional, the features are appended with the pseudo-negative data generated from a Gaussian, $ \mathcal{N}(\mu, \sigma \cdot {I})$, similar to \cite{oza2019one}. Following We use a simple one layer fully connected classifier network ($C$) with sigmoid activation at the end, as shown in Fig. \ref{fig:prop_app}. The number of hidden units are the same as the length of the feature vector representation. Because of the Gaussian vector concatenation at the input, the network $C$ observes twice the batch size ($N$) as of the feature extractor.

\subsection{Decoder Network}
The decoder network ($D$) architecture is a simple four layer fully convolutional neural network. This network takes feature representation learned by the network $E$ and tries to reconstruct the original input. This in effect constraints $E$ to generate representation which have self-representation property. It can be seen as a form of regularization on the feature representation. This regularization can be controlled with parameter $\lambda_r$ given in Eq. \ref{eq:loss_total} of total loss function. Since feature extractor outputs a flattened feature vector, we reshape it to an appropriate size before feeding it to the decoder network.  Note that $E$ along with $D$ can be viewed as an auto-encoder network. 

\subsection{Loss Functions}
The entire network is trained using a combination of two loss functions -  classification loss ($\mathcal{L}_c$) and reconstruction loss ($\mathcal{L}_r$). The classification loss is defined as follows

\begin{equation}
\mathcal{L}_c \ = \ - \frac{1}{2N} \sum_{j=1}^{2N} \ [ \ y \cdot \log_2(p) + (1-y) \cdot \log_2(1-p) \ ],
\label{eq:loss_bce}
\end{equation}

where  $y \in \{0, 1\}$ indicates whether classifier input corresponds to feature extractor (i.e., $y = 1$), or sampled from $ \mathcal{N}(\mu, \sigma \cdot {I})$, (i.e., $y = 0$). Here, $p$ is the probability of $y=1$. The classification network $C$ observes twice the input batch size because we append Gaussian vector in batch dimension with extracted features, in Eq. \ref{eq:loss_bce}, the summation is over $2N$.

The $L1$ reconstruction loss is defined as follows 

\begin{equation}
\mathcal{L}_r \ = \ \frac{1}{N} \sum_{j=1}^{N} \ \|X - \tilde{X}\|_1 
\label{eq:loss_l1},
\end{equation}

where $X$ and $\tilde{X}$ are the original input image and the reconstructed image, respectively.  

Finally, the overall loss function is the sum of $\mathcal{L}_r $ and $\mathcal{L}_c$  defined as follows

\begin{equation}
\mathcal{L}_t \ = \mathcal{L}_c + \lambda_r \mathcal{L}_r,
\label{eq:loss_total}
\end{equation}

where $\lambda_r$ is a regularization parameter.  Furthermore, note that $\tilde{X} = D(E(X))$ and $p = C(E(X))$.

The network is optimized using the Adam optimizer \cite{kingma2014adam} with the learning rate of $10^{-4}$ and batch size (i.e. $N$) of 64. For all the experiments, $\mu$, $\sigma$ and $\lambda_r$ are set equal to 0.0, 0.01 and 1.0, respectively. The decoder architecture is as follows\\
\noindent ConvTran(1024, 256) - ConvTran(256, 64) - ConvTran(64, 16) - ConvTran(16, 3),\\
where, ConvTran(\textit{in}, \textit{out}) denotes the transposed convolutions with \textit{in} and \textit{out} as number of input and output feature channels, respectively. All transposed convolutions are used with kernels of size $4 \times 4$. ReLU activation is used after every transposed convolution layer except the fourth, where Tanh activation is used. Instance normalization \cite{dumoulin2017learned} is used before the classifier network and at the end of every transposed convolution layer.

\begin{figure*}[!t]
    \centering
    \begin{subfigure}[t]{0.32\textwidth}
        \centering
        \includegraphics[height=1.5in]{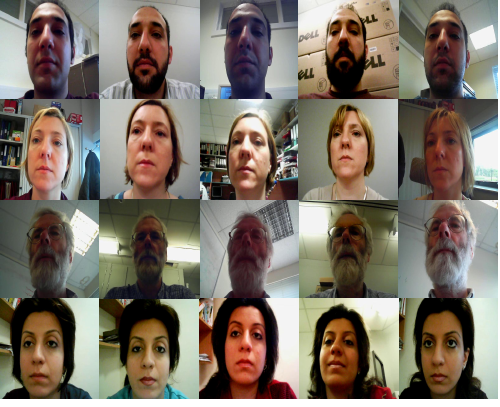}
        \caption{MOBIO}
        \label{fig:mobio}
    \end{subfigure}%
    ~ 
    \begin{subfigure}[t]{0.32\textwidth}
        \centering
        \includegraphics[height=1.5in]{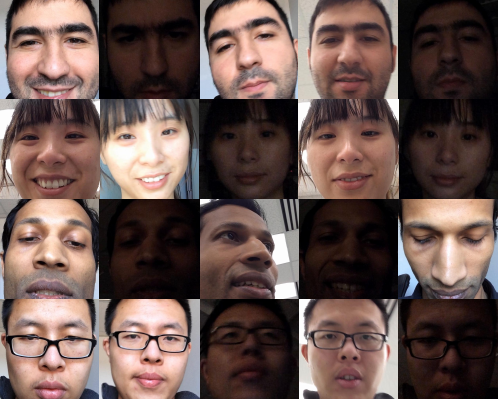}
        \caption{UMDAA-01 Face}
        \label{fig:umd1}
    \end{subfigure}
    ~
    \begin{subfigure}[t]{0.32\textwidth}
        \centering
        \includegraphics[height=1.5in]{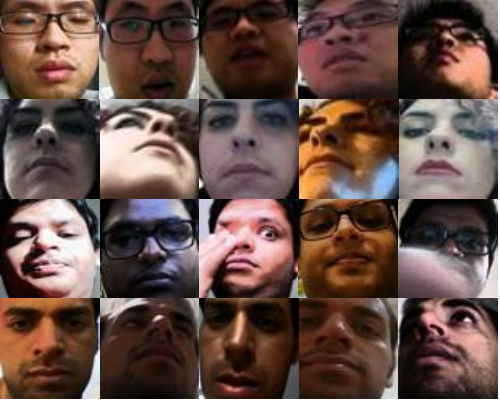}
        \caption{UMDAA-02 Face}
        \label{fig:umd2}
    \end{subfigure}    
    
   \caption{Sample Images from (a) Mobio, (b) UMDAA-01 Face, (c) UMDAA-02 Face datasets. Each column represents different user and each row shows multiple images from the same user.}
    \label{fig:sample_images}
\end{figure*}

\section{Experimental Results}
We evaluate the performance of the proposed approach on three publically available face-based AA datasets -- MOBIO \cite{tresadern2012mobile}, UMDAA-01 \cite{fathy2015face} and UMDAA-02 \cite{mahbub2016active}.  The proposed approach is compared with the following one-class classification methods:

\begin{enumerate}
	
	\item[$\bullet$] \textbf{OC-SVM:} One class SVM as formulated in \cite{scholkopf2001estimating} is used. OCSVM is trained on features extracted from AlexNet, VGG16 and VGGFace.
	
	\item[$\bullet$] \textbf{SMPM:} SMPM is used as formulated in \cite{lanckriet2002minimax}. In SMPM formulation, to utilize the second order statistics, covariance matrix computation is required. Hence, before applying SMPM, we reduce the dimensionality of the features extracted from AlexNet, VGG16 and VGGFace using Principle Component Analysis (PCA).
	
	\item[$\bullet$] \textbf{SVDD:} Support Vector Data Description is used as formulated in \cite{tax2004support}, trained on the AlexNet,VGG and VGGFace features.
	
	\item[$\bullet$] \textbf{OC-NN:} One-class neural network (OC-NN) is used as formulated in \cite{chalapathy2018anomaly}. The encoder network described in \cite{chalapathy2018anomaly} is replaced with a pretrained CNNs, i.e. AlexNet, VGG16 and VGGFace to have a fair comparison between the methods. Apart from this change, the training procedure is exactly the same as given in \cite{chalapathy2018anomaly}.
	
\end{enumerate}

The following ablation baselines are also considered to show the contribution of each module in the proposed approach:

\begin{enumerate}
	
	\item[$\bullet$] \textbf{Auto-Encoder baseline (only  $\mathcal{L}_r$):} This is one of the ablation baselines, where we utilize the feature extractor and the decoder networks, and train with only $\mathcal{L}_r$ loss function given in Eq. \ref{eq:loss_l1}. It can also be seen as a generative approach baseline. The reconstruction error is used for classification.  In other words, a pre-determined threshold is compared against the reconstruction error and the input is rejected if the error is greater than the threshold.  Otherwise, the input is declared as corresponding to the one-class data.  
	
	\item[$\bullet$] \textbf{Classifier baseline (only $\mathcal{L}_c$):} Another ablation baseline includes using classifier and feature extractor networks trained with only $\mathcal{L}_c$ loss function given in Eq. \ref{eq:loss_bce}. The classification network is not regularized by the decoder network. This baseline is equivalent to the method proposed in \cite{oza2019one}. This ablation study will clearly show the significance of using an auto-encoder as a  regularizer for one-class classification.
	
	\item[$\bullet$] \textbf{Proposed approach OC-ACNN (both  $\mathcal{L}_r$ and  $\mathcal{L}_r$):} OC-ACNN is the method proposed in this paper.
	
\end{enumerate}

For OC-SVM, SMPM and SVDD distance scores from the hyperplane/hypersphere are used for performance evaluation. For OC-NN, classifier baseline and the proposed approach, scores from the classifier are used for performance evaluation. As mentioned before, the reconstruction error is used for evaluating the performance of the auto-encoder baseline.

\subsection{Datasets}

\begin{figure*}[!t]
    \centering
    \begin{subfigure}[t]{0.23\textwidth}
        \centering
        \includegraphics[width=1.5in, height=1.5in]{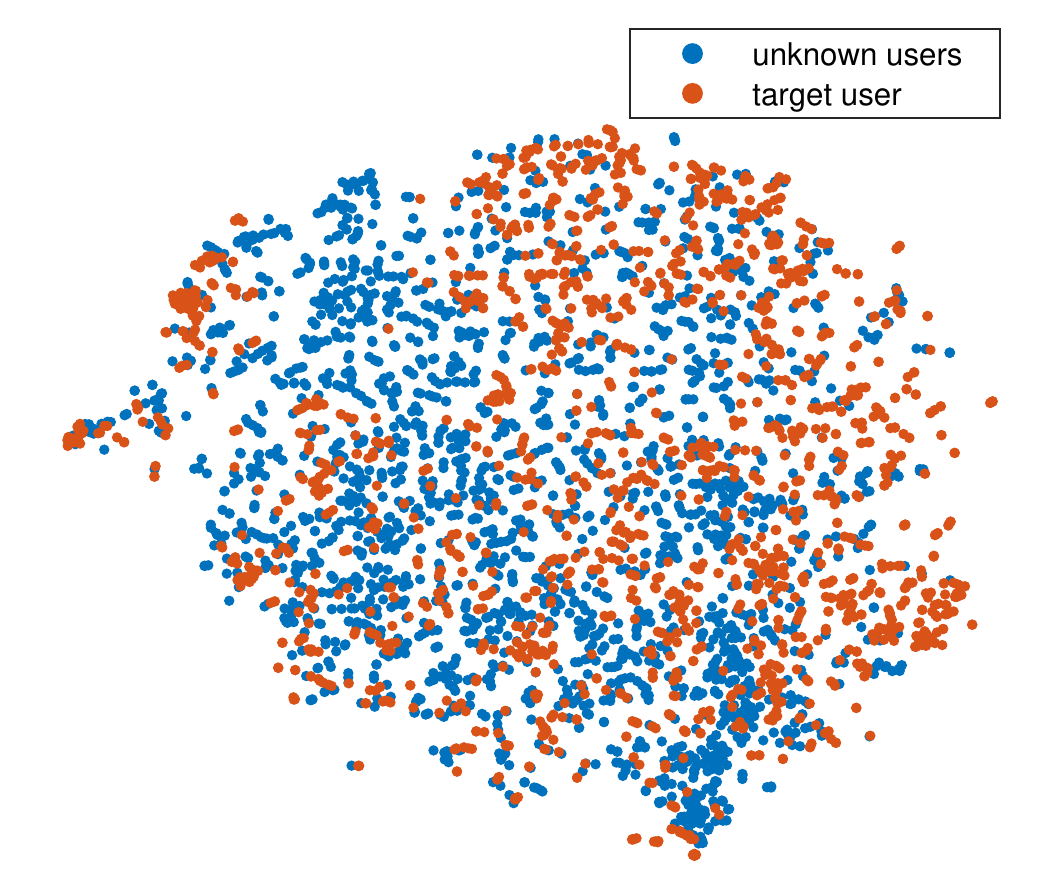}
        \caption{Pre-trained AlexNet.}
        \label{fig:pre_umd2_alexnet}
    \end{subfigure}%
    ~ 
    \begin{subfigure}[t]{0.23\textwidth}
        \centering
        \includegraphics[width=1.5in, height=1.5in]{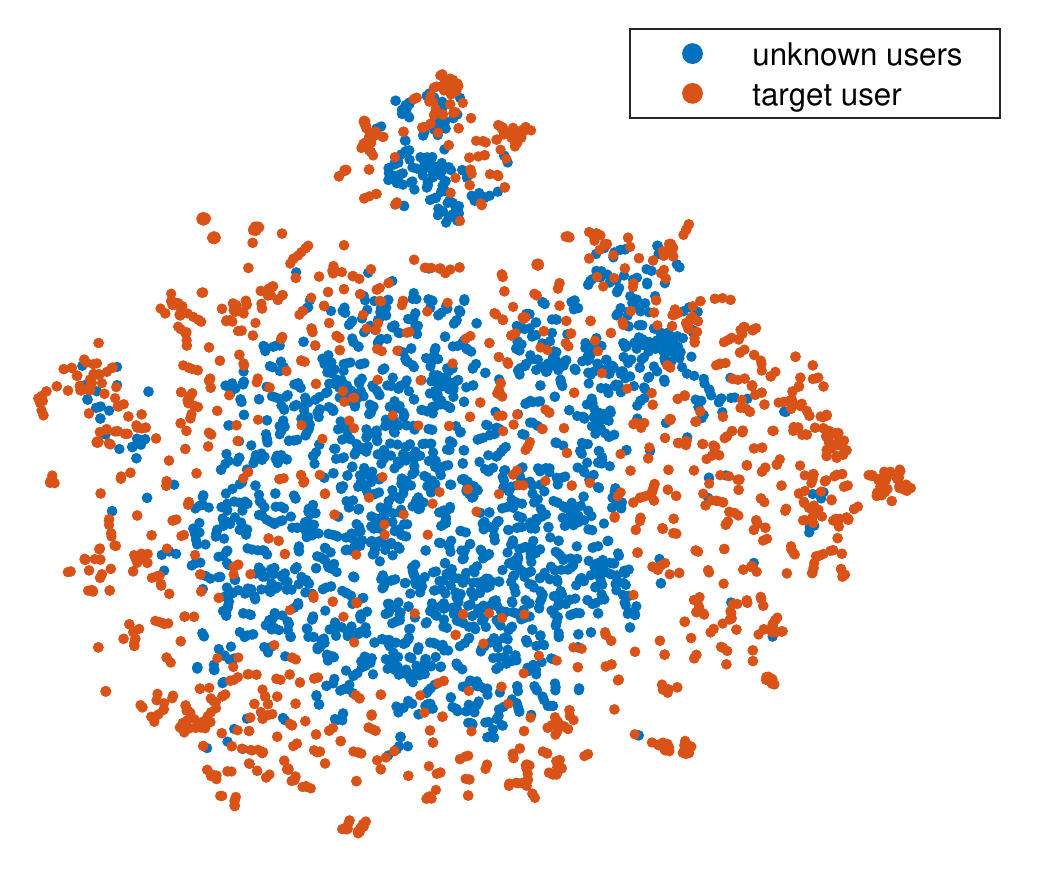}
        \caption{AlexNet trained using the propsoed method.}
        \label{fig:fin_umd2_alexnet}
    \end{subfigure}
    ~
    \begin{subfigure}[t]{0.23\textwidth}
        \centering
        \includegraphics[width=1.5in,height=1.5in]{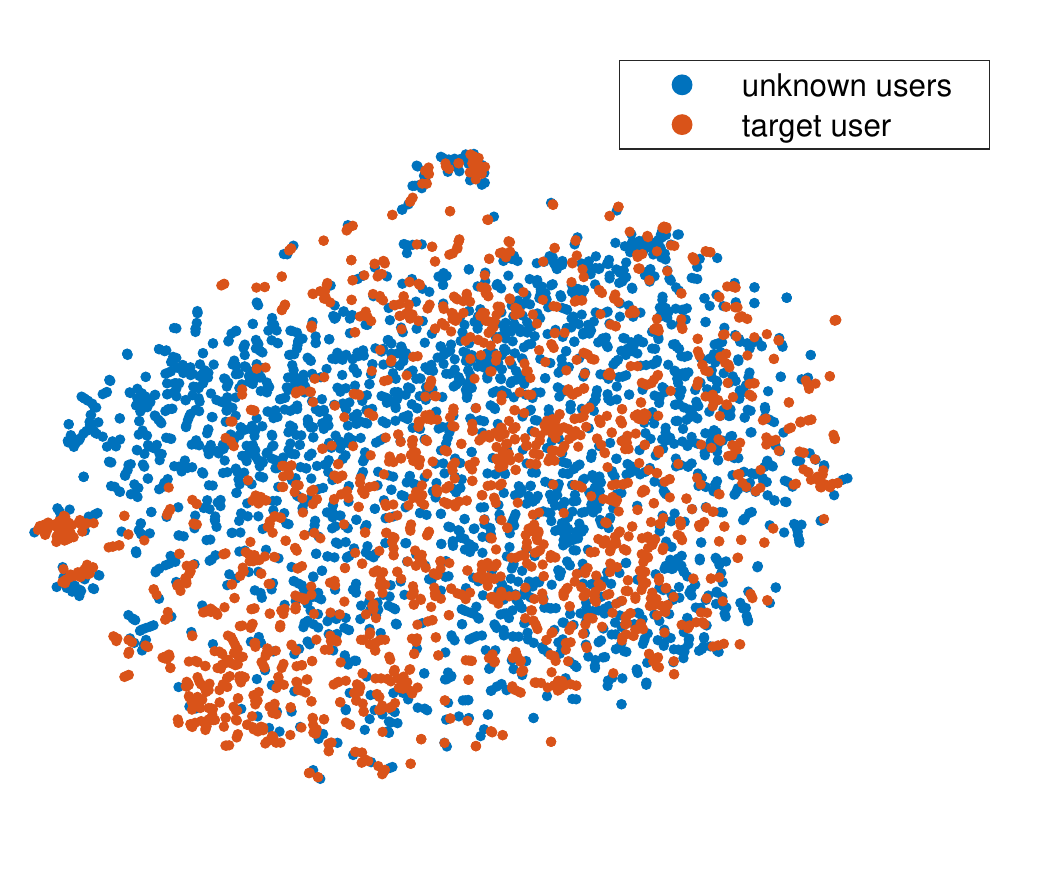}
        \caption{Pre-trained VGG16.}
        \label{fig:pre_umd2_vgg16}
    \end{subfigure}%
    ~ 
    \begin{subfigure}[t]{0.23\textwidth}
        \centering
        \includegraphics[width=1.5in,height=1.5in]{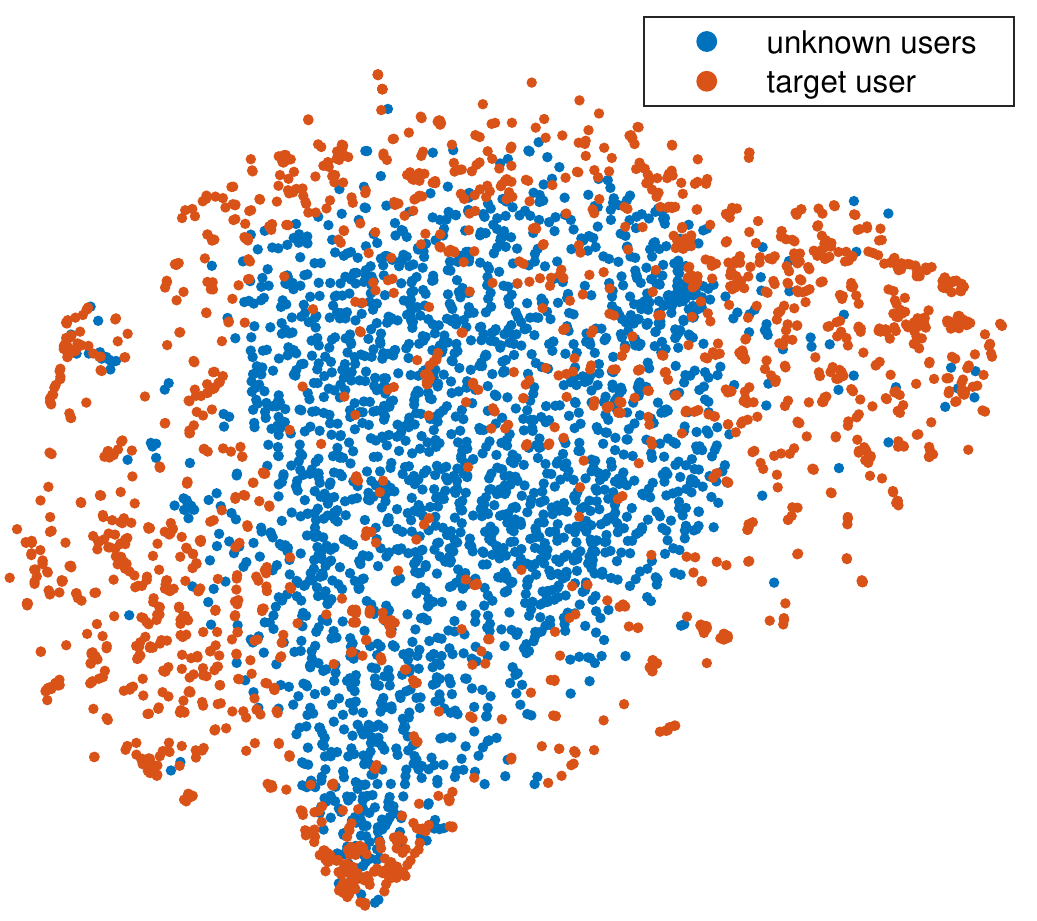}
        \caption{VGG16 trained using the propsoed method.}
        \label{fig:fin_umd2_vgg16}
    \end{subfigure}
     ~ 
    \begin{subfigure}[t]{0.23\textwidth}
        \centering
        \includegraphics[width=1.5in,height=1.5in]{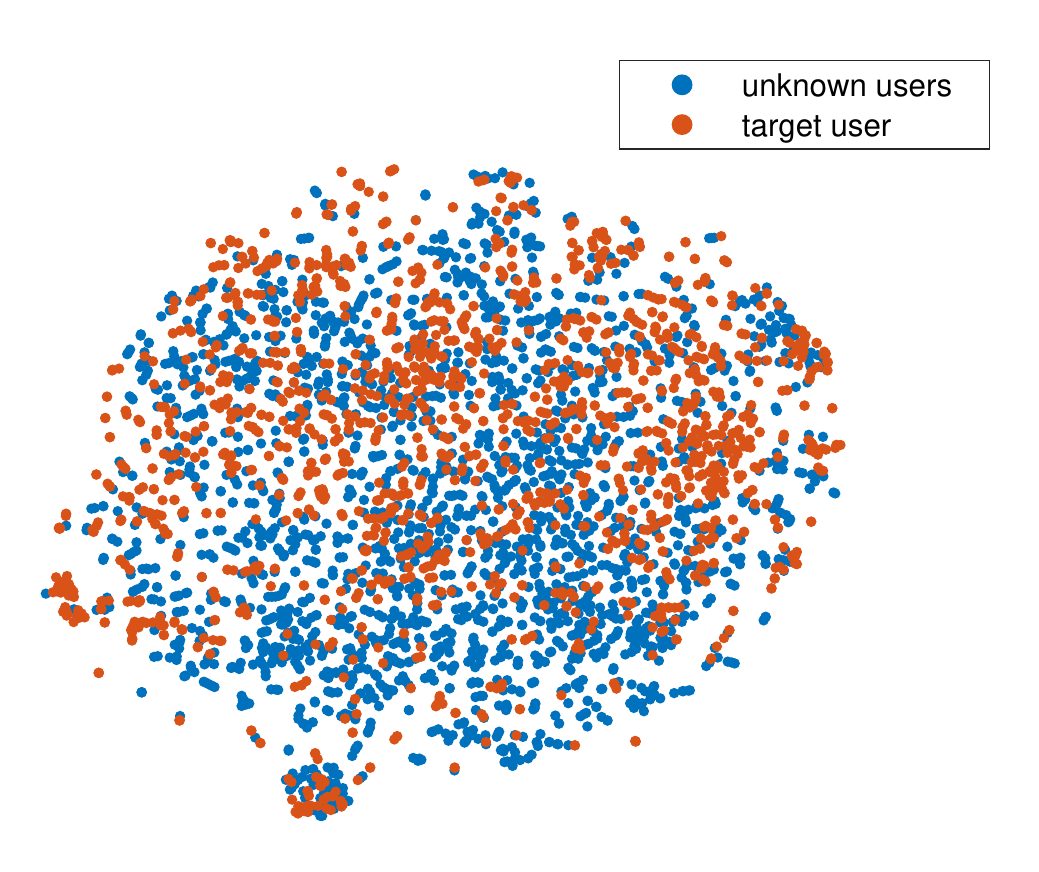}
        \caption{Pre-trained VGGFace.}
        \label{fig:pre_umd2_vggface}
    \end{subfigure}
    ~ 
    \begin{subfigure}[t]{0.23\textwidth}
        \centering
        \includegraphics[width=1.5in,height=1.5in]{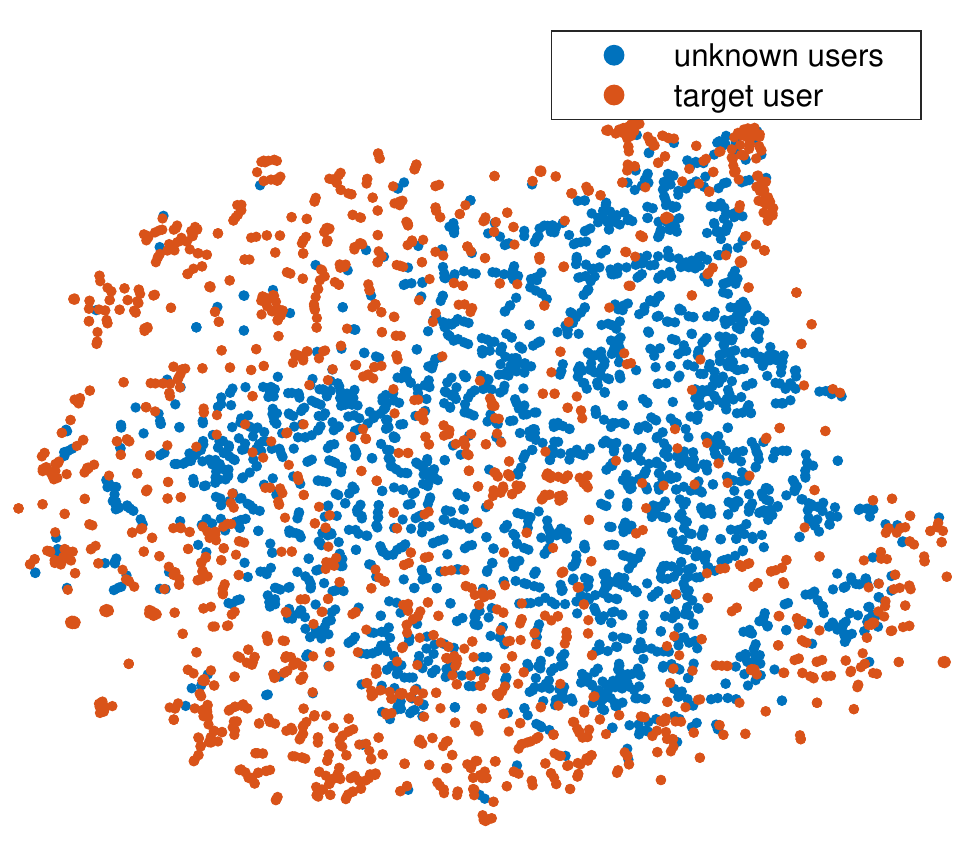}
        \caption{VGGFace trained using the propsoed method.}
        \label{fig:fin_umd2_vggface}
    \end{subfigure}
	   
    \caption{t-SNE visualizations of feature representations from the feature Extractor ($E$) corresponding to a user from the UMDAA-02 Face dataset.}
    \label{fig:qualitative_fig}
\end{figure*}

\textbf{MOBIO.} The MOBIO dataset is a bi-modal AA dataset containing face images and voice recordings of 150 users. In this paper, we only consider  face images for conducting the experiemnts. Sample images from this dataset are shown in Fig. \ref{fig:mobio}. For each user the recordings are taken in six sessions at different locations. We combine images from all six sessions. MOBIO contains less variations in pose, illumination etc., as compared to the other datasets used in this paper. For experiments, first 48 users are considered as target users and the rest are used as unknown users. Target users' data is split into train and test set with 85/15 ratio. For each target user, the training set is used to train the networks. During evaluation, we utilize the test set of the target user along with the data from all unknown users. This process is repeated for all 48 users and average performance is reported.\\

\textbf{UMDAA-01 Face.} The UMDAA-01 dataset contains face and touch gestures recorded from a mobile device. In total 750 video sequences of 50 users are collected in three different sessions with varying illumination conditions. Sample images from this dataset are shown in Fig. \ref{fig:umd1}. Data from different sessions are combined for each user and split into train and test sets with 80/20 ratio. Considering one user as the target and the remaining  49 users as unknown, networks are trained using target train set and tested with test set consisting of all 50 users' data (i.e. 1 target and 49 unknown). This experimental protocol is followed for all 50 users and average performance is reported.\\

\textbf{UMDAA-02 Face.} Unlike the above two datasets, the UMDAA-02 dataset has multiple modalities for 44 users e.g. face, gyroscope, swipe patterns, key strokes etc. all recorded from 18 sensors of a Nexus mobile device. Since the dataset was collected over a period of two months, it is an extremely challenging dataset with large variations in pose, illumination, occlusions and other environmental conditions. Fig. \ref{fig:umd2} shows some sample images from this dataset.  The number of sessions for each user ranges from 25 to 750 providing large number of frontal face images, i.e. more than 10k images on average per user. For each user, train and test splits are created with 80/20 ratio. We follow similar protocol as that of UMDAA01-Face for all 44 users and report the average performance.

Area Under the ROC curve (AUROC) is used to measure the performance. This is one of the most commonly used metric in the literature for evaluating the performance of the one-class classification methods.  

\subsection{Qualitative Evaluation}
In this section, we present qualitative evaluation of the proposed approach by comparing the visualizations of feature representations learned by our method with those corresponding to respective pre-trained networks. Fig. \ref{fig:qualitative_fig} shows t-SNE \cite{maaten2008visualizing} visualizations of the feature representations corresponding to AlexNet, VGG16 and VGGFace, respectively. These t-SNE plots are obtained from a single user of the UMDAA-02 Face dataset. Fig. \ref{fig:qualitative_fig}a, \ref{fig:qualitative_fig}c and \ref{fig:qualitative_fig}e show the visualizations corresponding to  pre-trained AlexNet, VGG16 and VGGFace networks, respectively.  Fig. \ref{fig:qualitative_fig}b, \ref{fig:qualitative_fig}d and \ref{fig:qualitative_fig}f show the visualizations corresponding to their counterpart networks trained using the proposed approach. As can be seen from these figures, the pre-trained networks generate features that highly overlap between the target and unknown users.  This makes sense since these networks are trained using a cross-entropy loss for multi-class classification. As a results, the features from these networks overlap significantly and it makes it difficult for a one-class classifier to correctly identify the separating decision boundary. On the other hand, the feature representations of the same networks trained using the proposed approach are quite distinctive. The learned feature representations corresponding to the target and unknown users are very well separated. These features become extremely useful while identifying the target user against the unknown users, thereby resulting in improved classification performance. 

Table \ref{table:tsne_table} shows the AUROC results corresponding to each plot for the same user computed using OC-SVM. As can be seen from this table, for all the networks the features learned using the proposed OC-ACNN provides better performance compared to the pre-trained features.  

\begin{table}[htp!]
	\centering
	\begin{tabular}{|c|c|c|}
		\hline
		\textbf{Feature Extractor ($E$)} & \textbf{Pre-Trained} & \textbf{OC-ACNN}  \\ \hline
		AlexNet & 0.5319  & 0.6780 \\ \hline
		VGG16   & 0.5698  & 0.8194 \\ \hline
		VGGFace & 0.5428  & 0.8808 \\ \hline
	\end{tabular}
	\caption{AUROC results corresponding to the study conducted in Fig.~\ref{fig:qualitative_fig}.}
	\label{table:tsne_table}
\end{table}

\begin{table*}[htbp]
	\centering
	\resizebox{\linewidth}{!}{%
	\begin{tabular}{|c|c|c|c|c|c|c|c|}
		\hline
		\textbf{Dataset} & \textbf{OC-SVM} & \textbf{SMPM} & \textbf{SVDD} & \textbf{OC-NN} & \textbf{Auto-encoder} (only $\mathcal{L}_r$) & \textbf{Classifier} (only $\mathcal{L}_c$) & \textbf{OC-ACNN} ($\mathcal{L}_c$+$\mathcal{L}_r$)           \\ \hline
		Mobio            & 0.6578 $\pm$ 0.1132  &  0.7721 $\pm$ 0.1185  & 0.7851 $\pm$ 0.1270 & 0.7504 $\pm$ 0.1512 & 0.7526 $\pm$ 0.1075 & 0.8191 $\pm$ 0.1286 & \textbf{0.8633 $\pm$ 0.1136}  \\ \hline
		UMDAA-01         & 0.6584 $\pm$ 0.1255 & 0.7576 $\pm$ 0.1149 & 0.8909 $\pm$ 0.0755 & 0.8684 $\pm$ 0.0913 & 0.6560 $\pm$ 0.1066 & 0.9196 $\pm$ 0.0482 & \textbf{0.9276  $\pm$ 0.0465}  \\ \hline
		UMDAA-02         & 0.5746 $\pm$ 0.0595 & 0.5418 $\pm$ 0.0382 & 0.6448 $\pm$ 0.0725 & 0.6542 $\pm$ 0.0593 & 0.5952 $\pm$ 0.0869 & 0.7017 $\pm$ 0.1007 & \textbf{0.7398 $\pm$ 0.0787}  \\ \hline
	\end{tabular}
}
	\caption{Comparison between the proposed approach and other one-class methods with \textbf{AlexNet} as the feature extractor network. Results are the mean of performance on all classes. The performance is measured by AUROC. Best performance is highlighted in bold fonts.}
	\label{table:auroc_alexnet}
\end{table*}

\begin{table*}[htbp]
	\centering
		\resizebox{\linewidth}{!}{%
	\begin{tabular}{|c|c|c|c|c|c|c|c|}
		\hline
		\textbf{Dataset} & \textbf{OC-SVM} & \textbf{SMPM} & \textbf{SVDD} & \textbf{OC-NN} & \textbf{Auto-encoder} (only $\mathcal{L}_r$) & \textbf{Classifier} (only $\mathcal{L}_c$) & \textbf{OC-ACNN} ($\mathcal{L}_c$+$\mathcal{L}_r$)           \\ \hline
		Mobio            & 0.6607 $\pm$ 0.1066 &  0.7266 $\pm$ 0.1046 & 0.8212 $\pm$ 0.1130 & 0.7822 $\pm$ 0.1153 & 0.7457 $\pm$ 0.1072 & 0.8177 $\pm$ 0.1132 & \textbf{0.8705 $\pm$ 0.1278}          \\ \hline
		UMDAA-01         & 0.6777 $\pm$ 0.0946 &  0.8664 $\pm$ 0.0765 & 0.9011 $\pm$ 0.0592 & 0.8802 $\pm$ 0.0976 & 0.8494 $\pm$ 0.0844 & 0.9348 $\pm$ 0.0384 & \textbf{0.9486 $\pm$ 0.0336}  \\ \hline
		UMDAA-02         & 0.5828 $\pm$ 0.0757 &  0.5473 $\pm$ 0.0447 & 0.6424 $\pm$ 0.0677 & 0.6199 $\pm$ 0.0693 & 0.6042 $\pm$ 0.0939 & 0.7349 $\pm$ 0.0845 & \textbf{0.8457 $\pm$ 0.0581}  \\ \hline
	\end{tabular}
}
	\caption{Comparison between the proposed approach and other one-class methods with \textbf{VGG16} as the feature extractor network. Results are the mean of performance on all classes. The performance is measured by AUROC. Best performance is highlighted in bold fonts.}
	\label{table:auroc_vgg16}
\end{table*}

\begin{table*}[htbp]
	\centering
		\resizebox{\linewidth}{!}{%
	\begin{tabular}{|c|c|c|c|c|c|c|c|}
		\hline
		\textbf{Dataset} & \textbf{OC-SVM} & \textbf{SMPM} & \textbf{SVDD} & \textbf{OC-NN} & \textbf{Auto-encoder} (only $\mathcal{L}_r$) & \textbf{Classifier} (only $\mathcal{L}_c$) & \textbf{OC-ACNN} ($\mathcal{L}_c$+$\mathcal{L}_r$)           \\ \hline
		Mobio            & 0.6702 $\pm$ 0.1268 &  0.6619 $\pm$ 0.1068  & 0.7975 $\pm$ 0.1250 & 0.7673 $\pm$ 0.1380 & 0.7339 $\pm$ 0.1095 & 0.8347 $\pm$ 0.1324 & \textbf{0.8859 $\pm$ 0.1042} \\ \hline
		UMDAA-01         & 0.6763 $\pm$ 0.1237 &  0.7334 $\pm$ 0.1241 & 0.8745 $\pm$ 0.0794 & 0.8257 $\pm$ 0.1381 & 0.8237 $\pm$ 0.0923 & 0.9432 $\pm$ 0.0654 & \textbf{0.9772 $\pm$ 0.0213} \\ \hline
		UMDAA-02         & 0.5712 $\pm$ 0.0644 &  0.5671 $\pm$ 0.0597 & 0.5898 $\pm$ 0.0647 & 0.5987 $\pm$ 0.0652 & 0.6343 $\pm$ 0.0723 & 0.6393 $\pm$ 0.0946 & \textbf{0.8946 $\pm$ 0.0535} \\ \hline
	\end{tabular}
}
	\caption{Comparison between the proposed approach and other one-class methods with \textbf{VGGFace} as the feature extractor network. Results are the mean of performance on all classes. The performance is measured by AUROC. Best performance is highlighted in bold fonts.}
	\label{table:auroc_vggface}
\end{table*}

\subsection{Quantitative Evaluation} 
Tables \ref{table:auroc_alexnet}, \ref{table:auroc_vgg16} and \ref{table:auroc_vggface} show the performance on all three datasets based on  AlexNet, VGG16 and VGGFace as feature extractors, respectively. The performance of other methods is inconsistent across the experiments. SMPM was found to perform better than OCSVM, while SVDD achieves better performance in many cases beating OC-NN. This may be due to the evaluation protocol difference between this paper and the one proposed in OC-NN \cite{chalapathy2018anomaly}. In OC-NN evaluation protocol, the number of unknown classes used during evaluation are much less than the number of unknown classes used for evaluation in this paper (i.e., MOBIO(96), UMDAA-01 Face(49) and UMDAA-02 Face(43)). This can be a reason for the poor performance from OC-NN as compared to SVDD. OC-NN however, manages to perform better than SMPM and OCSVM, and in couple of cases SVDD. Meanwhile the proposed approach achieves superior performance across all the datasets and for different feature extractor models.

Comparing the performance across models, VGGFace outperforms both VGG16 and AlexNet models.  This makes sense since face images (i.e. VGGFace dataset) were used to train the original VGGFace model and the corresponding weights are better suited for face-based AA application considered in this paper.  In contrast,  the VGG16 and AlexNet networks were trained using general object dataset (i.e. ImageNet dataset) for object recognition task.  The highest performance for all these networks is achieved for UMDAA-01 Face, since this dataset only contains illumination variations. Though MOBIO contains least variations in image samples, it has large number of unknown classes to compare against. While UMDAA-02 is the most difficult dataset since it contains very unconstrained face images.  As a result, the performance on this dataset is lower than the other two datasets. In summary, the proposed approach observes improvement of $\sim$6\%, $\sim$9\% and $\sim$16\% on average across all datasets corresponding to AlexNet, VGG16 and VGGFace, respectively.

Comparing ablation baselines, the auto-encoder baseline using only the reconstruction loss performs the poorest, while only the classifier baseline performs reasonably well. Auto-encoder and classifier baselines can be categorized as generative and discriminative approach, respectively.   Since the discriminative approach can learn better representation, it helps the classifier baseline to improve its performance. However, when the decoder is added to the classification pipeline to regularize the learned representations, it improves the overall performance by $\sim$6\%. This clearly shows the significance of enforcing the self-representation constraints to  regularize the learned feature representations for one-class classification.

\section{CONCLUSIONS}

We proposed a new approach for single user AA based on auto-encoder regularized CNNs. Feature representations are jointly learned with classifier influencing the generated representations. A pseudo-negative Gaussian vector was utilized to train the fully connected classification network. Decoder was introduced to regularize the generated feature representation by enforcing it to be self-representative. Experiments were conducted using the AlexNet, VGG16 and VGGFace networks, which showed the adaptability of the proposed method to work with different types of network architectures. Ablation study was conducted to show the importance of both classification loss and  feature regularization. Moreover, visualizations of the learned representations showed the ability of the proposed approach to learn distinctive features for one-class classification. Furthermore, the consistent performance improvements over all the datasets related to AA showed the significance of the proposed one-class classification method.

%


\bibliographystyle{IEEEtran}
\bibliography{refs}

\begin{thebibliography}{10}
\providecommand{\url}[1]{#1}
\csname url@samestyle\endcsname
\providecommand{\newblock}{\relax}
\providecommand{\bibinfo}[2]{#2}
\providecommand{\BIBentrySTDinterwordspacing}{\spaceskip=0pt\relax}
\providecommand{\BIBentryALTinterwordstretchfactor}{4}
\providecommand{\BIBentryALTinterwordspacing}{\spaceskip=\fontdimen2\font plus
\BIBentryALTinterwordstretchfactor\fontdimen3\font minus
  \fontdimen4\font\relax}
\providecommand{\BIBforeignlanguage}[2]{{%
\expandafter\ifx\csname l@#1\endcsname\relax
\typeout{** WARNING: IEEEtran.bst: No hyphenation pattern has been}%
\typeout{** loaded for the language `#1'. Using the pattern for}%
\typeout{** the default language instead.}%
\else
\language=\csname l@#1\endcsname
\fi
#2}}
\providecommand{\BIBdecl}{\relax}
\BIBdecl

\bibitem{prabhakar2003biometric}
S.~Prabhakar, S.~Pankanti, and A.~K. Jain, ``Biometric recognition: Security
  and privacy concerns,'' \emph{IEEE security \& privacy}, no.~2, pp. 33--42,
  2003.

\bibitem{cmaxine2016casestudy}
C.~M. Most, ``The global biometrics and mobility report: The convergence of
  commerce and privacy,'' \emph{Acuity Market Intelligence Report Received from
  : www.acuity-mi.com}, 2016.

\bibitem{guidorizzi2013security}
R.~P. Guidorizzi, ``Security: active authentication,'' \emph{IT Professional},
  vol.~15, no.~4, pp. 4--7, 2013.

\bibitem{pozo2017exploring}
A.~Pozo, J.~Fierrez, M.~Martinez-Diaz, J.~Galbally, and A.~Morales, ``Exploring
  a statistical method for touchscreen swipe biometrics,'' in \emph{Security
  Technology (ICCST), 2017 International Carnahan Conference on}.\hskip 1em
  plus 0.5em minus 0.4em\relax IEEE, 2017, pp. 1--4.

\bibitem{zhang2015touch}
H.~Zhang, V.~M. Patel, M.~Fathy, and R.~Chellappa, ``Touch gesture-based active
  user authentication using dictionaries,'' in \emph{2015 IEEE Winter
  Conference on Applications of Computer Vision (WACV)}.\hskip 1em plus 0.5em
  minus 0.4em\relax IEEE, 2015, pp. 207--214.

\bibitem{perera2017extreme}
P.~Perera and V.~M. Patel, ``Extreme value analysis for mobile active user
  authentication,'' in \emph{2017 12th IEEE International Conference on
  Automatic Face \& Gesture Recognition (FG 2017)}.\hskip 1em plus 0.5em minus
  0.4em\relax IEEE, 2017, pp. 346--353.

\bibitem{perera2017towards}
------, ``Towards multiple user active authentication in mobile devices,'' in
  \emph{Automatic Face \& Gesture Recognition (FG 2017), 2017 12th IEEE
  International Conference on}.\hskip 1em plus 0.5em minus 0.4em\relax IEEE,
  2017, pp. 354--361.

\bibitem{derawi2010unobtrusive}
M.~O. Derawi, C.~Nickel, P.~Bours, and C.~Busch, ``Unobtrusive
  user-authentication on mobile phones using biometric gait recognition,'' in
  \emph{Intelligent Information Hiding and Multimedia Signal Processing
  (IIH-MSP), 2010 Sixth International Conference on}.\hskip 1em plus 0.5em
  minus 0.4em\relax IEEE, 2010, pp. 306--311.

\bibitem{fathy2015face}
M.~E. Fathy, V.~M. Patel, and R.~Chellappa, ``Face-based active authentication
  on mobile devices,'' in \emph{Acoustics, Speech and Signal Processing
  (ICASSP), 2015 IEEE International Conference on}.\hskip 1em plus 0.5em minus
  0.4em\relax IEEE, 2015, pp. 1687--1691.

\bibitem{crouse2015continuous}
D.~Crouse, H.~Han, D.~Chandra, B.~Barbello, and A.~K. Jain, ``Continuous
  authentication of mobile user: Fusion of face image and inertial measurement
  unit data,'' in \emph{Biometrics (ICB), 2015 International Conference
  on}.\hskip 1em plus 0.5em minus 0.4em\relax IEEE, 2015, pp. 135--142.

\bibitem{samangouei2016convolutional}
P.~Samangouei and R.~Chellappa, ``Convolutional neural networks for
  attribute-based active authentication on mobile devices,'' in
  \emph{Biometrics Theory, Applications and Systems (BTAS), 2016 IEEE 8th
  International Conference on}.\hskip 1em plus 0.5em minus 0.4em\relax IEEE,
  2016, pp. 1--8.

\bibitem{perera2018efficient}
P.~Perera and V.~M. Patel, ``Efficient and low latency detection of intruders
  in mobile active authentication,'' \emph{IEEE Transactions on Information
  Forensics and Security}, vol.~13, no.~6, pp. 1392--1405, 2018.

\bibitem{perera2016quickest}
------, ``Quickest intrusion detection in mobile active user authentication,''
  in \emph{2016 IEEE 8th International Conference on Biometrics Theory,
  Applications and Systems (BTAS)}.\hskip 1em plus 0.5em minus 0.4em\relax
  IEEE, 2016, pp. 1--8.

\bibitem{patel2016continuous}
V.~M. Patel, R.~Chellappa, D.~Chandra, and B.~Barbello, ``Continuous user
  authentication on mobile devices: Recent progress and remaining challenges,''
  \emph{IEEE Signal Processing Magazine}, vol.~33, no.~4, pp. 49--61, 2016.

\bibitem{antal2015evaluation}
M.~Antal and L.~Z. Szab{\'o}, ``An evaluation of one-class and two-class
  classification algorithms for keystroke dynamics authentication on mobile
  devices,'' in \emph{Control Systems and Computer Science (CSCS), 2015 20th
  International Conference on}.\hskip 1em plus 0.5em minus 0.4em\relax IEEE,
  2015, pp. 343--350.

\bibitem{scholkopf2001estimating}
B.~Sch{\"o}lkopf, J.~C. Platt, J.~Shawe-Taylor, A.~J. Smola, and R.~C.
  Williamson, ``Estimating the support of a high-dimensional distribution,''
  \emph{Neural computation}, vol.~13, no.~7, pp. 1443--1471, 2001.

\bibitem{tax2004support}
D.~M. Tax and R.~P. Duin, ``Support vector data description,'' \emph{Machine
  learning}, vol.~54, no.~1, pp. 45--66, 2004.

\bibitem{ghaoui2003robust}
L.~E. Ghaoui, M.~I. Jordan, and G.~R. Lanckriet, ``Robust novelty detection
  with single-class mpm,'' in \emph{Advances in neural information processing
  systems}, 2003, pp. 929--936.

\bibitem{khan2014one}
S.~S. Khan and M.~G. Madden, ``One-class classification: taxonomy of study and
  review of techniques,'' \emph{The Knowledge Engineering Review}, vol.~29,
  no.~3, pp. 345--374, 2014.

\bibitem{goodfellow2014generative}
I.~Goodfellow, J.~Pouget-Abadie, M.~Mirza, B.~Xu, D.~Warde-Farley, S.~Ozair,
  A.~Courville, and Y.~Bengio, ``Generative adversarial nets,'' in
  \emph{Advances in neural information processing systems}, 2014, pp.
  2672--2680.

\bibitem{sabokrou2018adversarially}
M.~Sabokrou, M.~Khalooei, M.~Fathy, and E.~Adeli, ``Adversarially learned
  one-class classifier for novelty detection,'' in \emph{Proceedings of the
  IEEE Conference on Computer Vision and Pattern Recognition}, 2018, pp.
  3379--3388.

\bibitem{sabokrou2018deep}
M.~Sabokrou, M.~Fayyaz, M.~Fathy, Z.~Moayed, and R.~Klette, ``Deep-anomaly:
  Fully convolutional neural network for fast anomaly detection in crowded
  scenes,'' \emph{Computer Vision and Image Understanding}, 2018.

\bibitem{lawson2017finding}
W.~Lawson, E.~Bekele, and K.~Sullivan, ``Finding anomalies with generative
  adversarial networks for a patrolbot,'' in \emph{Proceedings of the IEEE
  Conference on Computer Vision and Pattern Recognition Workshops}, 2017, pp.
  12--13.

\bibitem{ravanbakhsh2017abnormal}
M.~Ravanbakhsh, M.~Nabi, E.~Sangineto, L.~Marcenaro, C.~Regazzoni, and N.~Sebe,
  ``Abnormal event detection in videos using generative adversarial nets,'' in
  \emph{Image Processing (ICIP), 2017 IEEE International Conference on}.\hskip
  1em plus 0.5em minus 0.4em\relax IEEE, 2017, pp. 1577--1581.

\bibitem{zhou2017anomaly}
C.~Zhou and R.~C. Paffenroth, ``Anomaly detection with robust deep
  autoencoders,'' in \emph{Proceedings of the 23rd ACM SIGKDD International
  Conference on Knowledge Discovery and Data Mining}.\hskip 1em plus 0.5em
  minus 0.4em\relax ACM, 2017, pp. 665--674.

\bibitem{erfani2016high}
S.~M. Erfani, S.~Rajasegarar, S.~Karunasekera, and C.~Leckie,
  ``High-dimensional and large-scale anomaly detection using a linear one-class
  svm with deep learning,'' \emph{Pattern Recognition}, vol.~58, pp. 121--134,
  2016.

\bibitem{andrews2016transfer}
J.~T. Andrews, ``Transfer representation-learning for anomaly detection.''

\bibitem{perera2018learning}
P.~Perera and V.~M. Patel, ``Learning deep features for one-class
  classification,'' \emph{arXiv preprint arXiv:1801.05365}, 2018.

\bibitem{chalapathy2018anomaly}
R.~Chalapathy, A.~K. Menon, and S.~Chawla, ``Anomaly detection using one-class
  neural networks,'' \emph{arXiv preprint arXiv:1802.06360}, 2018.

\bibitem{lanckriet2002minimax}
G.~Lanckriet, L.~E. Ghaoui, C.~Bhattacharyya, and M.~I. Jordan, ``Minimax
  probability machine,'' in \emph{Advances in neural information processing
  systems}, 2002, pp. 801--807.

\bibitem{oza2019one}
P.~Oza and V.~M. Patel, ``One-class convolutional neural network,'' \emph{IEEE
  Signal Processing Letters}, vol.~26, no.~2, pp. 277--281, 2019.

\bibitem{krizhevsky2012imagenet}
A.~Krizhevsky, I.~Sutskever, and G.~E. Hinton, ``Imagenet classification with
  deep convolutional neural networks,'' in \emph{Advances in neural information
  processing systems}, 2012, pp. 1097--1105.

\bibitem{simonyan2014very}
K.~Simonyan and A.~Zisserman, ``Very deep convolutional networks for
  large-scale image recognition,'' 2015.

\bibitem{parkhi2015deep}
O.~M. Parkhi, A.~Vedaldi, A.~Zisserman \emph{et~al.}, ``Deep face
  recognition.'' in \emph{BMVC}, vol.~1, no.~3, 2015, p.~6.

\bibitem{kingma2014adam}
D.~P. Kingma and J.~Ba, ``Adam: A method for stochastic optimization,''
  \emph{arXiv preprint arXiv:1412.6980}, 2014.

\bibitem{dumoulin2017learned}
V.~Dumoulin, J.~Shlens, and M.~Kudlur, ``A learned representation for artistic
  style,'' \emph{Proc. of ICLR}, 2017.

\bibitem{tresadern2012mobile}
P.~Tresadern, C.~McCool, N.~Poh, P.~Matejka, A.~Hadid, C.~Levy, T.~Cootes, and
  S.~Marcel, ``Mobile biometrics (mobio): Joint face and voice verification for
  a mobile platform,'' \emph{IEEE pervasive computing}, 2012.

\bibitem{mahbub2016active}
U.~Mahbub, S.~Sarkar, V.~M. Patel, and R.~Chellappa, ``Active user
  authentication for smartphones: A challenge data set and benchmark results,''
  in \emph{Biometrics Theory, Applications and Systems (BTAS), 2016 IEEE 8th
  International Conference on}.\hskip 1em plus 0.5em minus 0.4em\relax IEEE,
  2016, pp. 1--8.

\bibitem{maaten2008visualizing}
L.~v.~d. Maaten and G.~Hinton, ``Visualizing data using t-sne,'' \emph{Journal
  of machine learning research}, vol.~9, no. Nov, pp. 2579--2605, 2008.

\end{thebibliography}

%
%
%
%

\end{document}